# Generalizable Pancreas Segmentation via a Dual Self-Supervised Learning Framework

Jun Li, Hongzhang Zhu, Tao Chen, and Xiaohua Qian

*Abstract*—Recently, numerous pancreas segmentation methods have achieved promising performance on local single-source datasets. However, these methods don't adequately account for generalizability issues, and hence typically show limited performance and low stability on test data from other sources. Considering the limited availability of distinct data sources, we seek to improve the generalization performance of a pancreas segmentation model trained with a single-source dataset, i.e., the single-source generalization task. In particular, we propose a dual self-supervised learning model that incorporates both global and local anatomical contexts. Our model aims to fully exploit the anatomical features of the intra-pancreatic and extra-pancreatic regions, and hence enhance the characterization of the high-uncertainty regions for more robust generalization. Specifically, we first construct a global-feature contrastive self-supervised learning module that is guided by the pancreatic spatial structure. This module obtains complete and consistent pancreatic features through promoting intra-class cohesion, and also extracts more discriminative features for differentiating between pancreatic and non-pancreatic tissues through maximizing inter-class separation. It mitigates the influence of surrounding tissue on the segmentation outcomes in high-uncertainty regions. Subsequently, a local-image-restoration self-supervised learning module is introduced to further enhance the characterization of the high-uncertainty regions. In this module, informative anatomical contexts are actually learned to recover randomly-corrupted appearance patterns in those regions. The effectiveness of our method is demonstrated with state-of-the-art performance and comprehensive ablation analysis on three pancreas datasets (467 cases). The results demonstrate a great potential in providing a stable support for the diagnosis and treatment of pancreatic diseases.

## I. INTRODUCTION

PANCREATIC cancer is an aggressive malignancy with one of the highest mortality rates, where the most effective treatment is early surgical resection combined with post‑operative chemotherapy. Accurate pancreas segmentation in CT images plays important roles in the diagnosis and treatment of this malignancy. Indeed, as pancreas segmentation typically produces accurate pancreas contours, it helps with surgical and radiotherapy planning, and tumor progression monitoring [1-3]. However, the pancreas is considered one of the most challenging abdominal organs to segment in CT images due to its obscure boundaries, drastically-changing shape, and small volume. Numerous studies [4-6] sought to overcome these challenges using convolutional neural networks (CNNs), and promising results were reported on local datasets. However, CNNs are data-driven models for which data source variability can cause a clear mismatch between the training data and the testing data, and can thus lead to a significant degradation in segmentation performance on unseen data. To the best of our knowledge, no independent testing on external data has been performed for any of the existing CNN-based pancreas segmentation methods. Thus, further exploration of the generalization capabilities of these methods is still required to assess their applicability in broader clinical contexts.

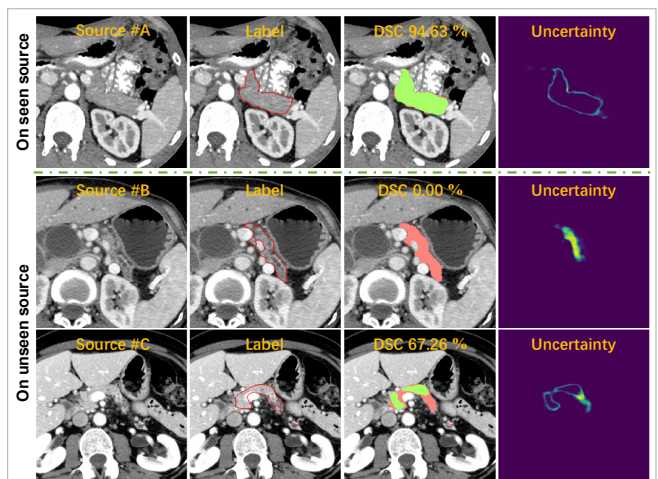

Fig. 1. The performance gap between seen and unseen sources. The first row shows robust performance on seen sources, while the second and third rows show a dramatic performance drop on unseen sources.

In computer-aided medical decision-support systems, the performance of a model trained on data obtained from one specific medical center usually degrades when applied to testing data from other centers [7]. This can be attributed to the

This work was supported by the Natural Science Foundation of Shanghai under Grant 22ZR1432100, National Natural Science Foundation of China under Grant 62171273, and Med-Engineering Crossing Foundation from Shanghai Jiao Tong University under Grant YG2022QN007. *(J. Li and H. Zhu contributed equally) (Corresponding author: T. Chen; X. Qian).*

J. Li and X. Qian are with the Medical Image and Health Informatics Lab, School of Biomedical Engineering, Shanghai Jiao

Tong University, Shanghai 200030, China (e-mail: dirk_li@sjtu.edu.cn; xiaohua.qian@sjtu.edu.cn).

H. Zhu is with Department of Radiology, The First Affiliated Hospital of Sun Yat-Sen University, Guangzhou, 510080, China (e-mail: zhhzhang@mail.sysu.edu.cn).

T. Chen is with Department of Biliary-Pancreatic Surgery, Renji Hospital, School of Medicine, Shanghai Jiao Tong University, Shanghai, China (e-mail: dr_chentao78@163.com).

premise that high CNN performance typically relies on having training and testing data from similar statistical distributions [8]. That is, the model can perform consistently on training and testing data when the premise is held. Unfortunately, this assumption fails to hold in practical settings due to various image acquisition factors, which lead to wide variations in intensity statistics, grayscale texture and contrast, etc. The appearance discrepancy among images from different sources leads to a significant decrease in the segmentation accuracy and stability in the generalization task (as shown in Fig. 1).

To address this issue, a common practice is to collect training data from the target environment before model deployment, where all source and target samples are paired and labelled, to fine-tune the model for better performance on target samples. Although this fine-tuning strategy can be effective, it is hard to extend this strategy to a wider range of application scenarios due to the significant challenges of collecting large datasets in advance. Another strategy is to aggregate data from multiple sources to increase the real-world variability within the data. This drives adaptation to potential imaging discrepancies for better generalization [9]. However, such a strategy is usually hindered by the non-availability of multi-source datasets, possibly because of the high collection and annotation costs and other challenges of multicenter studies [10]. Thus, a robust single-source model can be a viable and practical alternative to compensate for the drop in generalization performance when multi-source datasets are not available. In this work, we focus on this alternative of employing a robust model for boosting the generalization performance and robustness on unseen data when only one source is actually available for training.

Most single-source generalization methods [8, 11-13] aimed to enhance the adaptability through training data perturbation, which could be constrained by the limited amount of single-source data. Thus, we seek to adequately employ the anatomical context, which is a unique point in medical images, to enhance the generalization. Our goal is to fully exploit anatomical structural features to alleviate the pancreas segmentation challenges posed by the limited amount of single-source data and imaging discrepancies. Recently, [14-16] demonstrated a potential correlation between presence of high uncertainties and performance degradation in some challenging tasks. On the one hand, the pancreatic body is usually shown to be associated with a lower uncertainty and greater robustness, mainly attributed to the relatively high contrast between the pancreatic body and its surroundings and also attributed to rich anatomical context. On the other hand, high uncertainties are mostly observed in the head and tail of the pancreas or in other regions with drastic shape changes, essentially due to blurred boundaries and poor anatomical context. Thus, the segmentation performance is generally degraded by the presence of high-uncertainty regions, and the wide imaging variabilities in unseen test data. Fig. 1 demonstrates that the imaging variations cause a significant performance drop in high-uncertainty regions. Consequently, it is highly desirable to pay special attention to the high-uncertainty regions to enhance their characterization and boost the performance in the presence of these regions.

Thus, we consider two strategies based on the pancreatic external and internal anatomical environments, to enhance the performance within the challenging high-uncertainty regions.

*a) Mitigating the influence of the surrounding tissues to obtain stable representations of high-uncertainty regions:* We realized that poor pancreas segmentation is associated with high confusion with its surroundings, especially in generalization tasks where the appearance is highly variable. Thus, reducing the interference of these surrounding tissues is crucial for obtaining robust pancreatic feature representations. One key approach towards that goal is contrastive learning which has gained wide interest in medical image community. Indeed, it can lead to results with accurate boundaries through jointly maximizing intra-class cohesion and inter-class separation. However, one issue that needs to be explored is that contrastive learning is often applied in weakly-supervised or unsupervised learning tasks but not in supervised learning tasks [17, 18].

*b) Exploiting the spatial continuity of the pancreatic anatomical structures to strengthen the characterization of the high-uncertainty regions*: The exploitation of neighborhood information is widely recognized to be beneficial in object detection and segmentation tasks [19, 20]. Many studies [21, 22] have demonstrated successful restoration of diverse medical images, encompassing varied modalities and body parts, such as COVID CT, brain MRI, abdominal CT, and abdominal MRI. It demonstrates that the damaged small regions can be essentially recovered with the semantic information from the surrounding regions. In fact, accurate feature representations can be acquired based on the anatomical context and can facilitate other medical image analysis tasks [23, 24]. Thus, considering the visual and spatial continuity and anatomical context within the pancreatic tissues, we apply restorative self-supervised learning to improve the characterization of the high-uncertainty regions.

Motivated by the above observations, we propose a dual self-supervised generalization model, which exploits global and local clues and fully explores anatomical structural features, to enhance the characterization of the high-uncertainty regions for robust pancreas segmentation. Specifically, we do the following:

*Solution-a) We construct a global-feature self-supervised contrastive learning module using intra-pancreatic and extra-pancreatic feature embeddings.* It mitigates the influence of the extra-pancreatic tissues on the representation of the high-uncertainty regions. The contrastive learning pushes away the features of interfering regions, leading to more discriminative representations. To ensure the representativeness, completeness, and consistency of the intra-pancreatic feature embeddings, we scan the whole pancreatic region along its centerline rather than focusing only on the border contours [25, 26].

*Solution-b) We design a local-image-restoration self-supervised learning module with a multi-task collaboration capability.* It exploits the anatomical context to get enhanced characterization of the high-uncertainty regions. This is achieved through an image-restoration task where high-uncertainty regions are subjected to random appearance defects, and the appearance is restored in an adversarial manner. Self-supervised learning is employed with pixel-level supervision and local anatomical context to achieve better recovery and characterization of the high-uncertainty regions. Thus, the enhanced characterization of the high-uncertainty regions is indeed beneficial for boosting the generalization performance in pancreas segmentation.

The main contributions are summarized as follows:

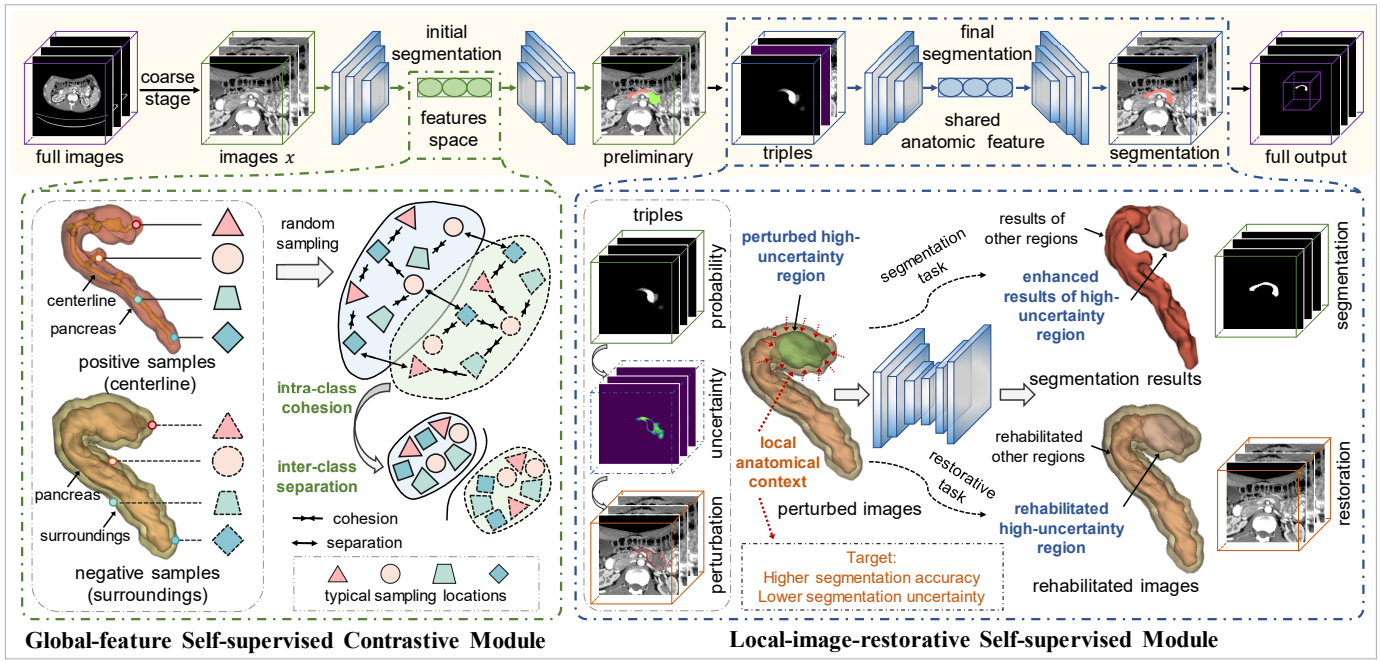

Fig. 2. An illustrated workflow of dual self-supervised learning model. It follows an end-to-end coarse-to-fine workflow. The pancreas is localized in coarse stage, then the proposed method is applied in initial and fine segmentation to finally output the result with the same size as inputs.

1) A dual self-supervised framework for systematically mining rich anatomical information in pancreatic imaging. This framework enhances the stability and accuracy of the characterization of the high-uncertainty regions and enables more robust single-source generalization performance.

2) A global self-supervised contrastive learning strategy for mitigating the influence of the pancreatic surroundings on sensitive high-uncertainty regions. In this strategy, the global feature boundary between the pancreas and its surroundings is expanded, leading to more stable feature representations.

3) A local-image-restoration self-supervised learning strategy for mining the local anatomical context, recovering the visual appearance in high-uncertainty regions, and enhancing the characterization of these sensitive regions.

## II. RELATED WORKS

### A. Pancreas Segmentation Methods

Existing pancreas segmentation methods can be generally classified into two categories according to their workflow, namely one-stage methods [6, 27-29] and two-stage methods [4, 5, 30-34]. The one-stage method aims to obtain results in one step. For example, Khosravan et al. [6] developed a spatial projection strategy based on 2D convolution for 3D information processing. Fang et al. [28] used a globally-guided progressive network based on 2D and 3D feature fusion. Zhang et al. [27] proposed a method based on nn-UNet [35] where a sliding window is applied to divide the raw image into 3D patches, and then the results of these patches are assembled to obtain the final output. The two-stage approach locates an approximate region of interest (ROI) by coarse segmentation, and then performs fine segmentation within the ROI. Zhu et al. [4] used two 3D networks for coarse and fine segmentation. Roth et al. [30] employed three 2D networks to aggregate spatial information from the coronal, sagittal and axial views. The seminal work in [30] was immediately followed by essentially similar variants [5, 31-33]. Since the two-stage models can focus on their respective tasks in different stages, these models have been more frequently employed to alleviate the challenges of the global search for the best pancreas segmentation maps.

### B. Single-source Generalization Methods

The single-source generalization task is highly challenging as it relies on only single-source data to get stable results. Most methods aim to generate assortments of appearances using data augmentation or synthesis methods [8, 11-13]. For example, [8] and [36] applied perturbation techniques to create rich image variations, and improve the model adaptability to such variations in real images. In [37, 38], feature perturbation was performed, maximum-gradient features were removed, new styles with mixed abstract features were generated, and hence the model adaptability was improved. However, the generated style variations are limited by the number of samples of the single-source data. [12] and [13] applied adversarial enhancement to synthetic data. But synthesizing volumetric data is quite harder than that of synthesizing 2D images, while slice-by-slice synthesis may damage the spatial anatomical information. Thus, effective synthesis of 3D medical image remains a key challenge. Karani et al. [39] proposed a test-time adaptation strategy. However, re-adapting each sample significantly increases the overhead, thus undermines the efficiency. Thus, we aim to fully exploit the anatomical context for enhancing the characterization of the high-uncertainty regions, thus improving the single-source generalization segmentation performance.

## III. METHODOLOGY

Fig. 2 illustrates the workflow, a coarse-to-fine strategy is used, and our method is realized within the fine stage. A global-feature self-supervised contrastive learning module eliminates the interference from surroundings. Local-image-restoration self-supervised learning module exploits the anatomical context of high-uncertainty regions, better characterize these regions, hence boost the segmentation. Our method considers both global and local information, promotes full exploitation of the anatomical structural features of each individual.

### A. Global-feature Anatomical-structure-guided Self-supervised Contrastive Learning

Earlier contrastive methods usually constructed inter-image instances pairs and overlooked the role of intra-image information, especially for medical data with valuable anatomical information. To alleviate the influence of the tissues surrounding the pancreas, we employ a global-feature contrastive learning module to cluster the pancreatic and non-pancreatic features, and hence set clear distinctive boundaries between the features of the two classes. Specifically, we extract the output $f_{out} \in \mathbb{R}^{B \times C \times D \times H \times W}$, where $B$ denotes the batch size and is set as 1, $C$ denotes the number of feature channels which is set as 256. The dimensions $D, H, W$ represent the slices, height, and width of the feature maps, respectively. These dimensions may vary for different samples due to the variation in pancreatic sizes. Then, a projection head is used to further introduce nonlinearity and reduce the channel dimension to 32 for $f_{proj} \in \mathbb{R}^{B \times c \times D \times H \times W}$, where $c$ is 32. Due to the batchsize being 1, $B$ will be omitted in subsequent presentations. Here, we use a 1×1 convolution as the projection head. Sub-patches are selected from $f_{proj}$ to construct positive and negative pairs, with optimizations for maximizing intra-class cohesion and inter-class separation.

#### 1) Constructing positive samples

The pancreatic anatomy is spatially continuous, physicians often delineate the pancreas by viewing adjacent slices. Thus, a promising segmentation model should approximate representations for different pancreatic parts (e.g., the head, body, and tail) instead of focusing only on the borders. However, the models tend to output unstable results with high uncertainties due to the dramatically-changing shape and imaging discrepancies. These typically lead to incomplete or biased representations. Thus, we introduced the anatomical information as a priori knowledge for contrastive learning. By collecting positive samples along the centerline, the features obtained from the head to the tail regions are characterized by high consistency and coherence. This leads to complete and representative representations. The process of selecting positive samples can be divided into two distinctive steps. Initially, a 3D medial surface/axis thinning algorithm [40] is employed to derive the complete centerline of the pancreas from the label, followed by a dilation operation to generate a mask. The spatial relationship between the mask and the feature maps enables us to guarantee that each positive sample is extracted from the centerline mask region. Subsequently, the feature map $f_{proj} \in \mathbb{R}^{c \times D \times H \times W}$ is traversed slice by slice, and a sub-patch is randomly selected from the centerline mask region of each slice in feature map $f_{proj}$ to create a positive sample $f_p^{sli} \in \mathbb{R}^{c \times h \times w}$, where $h \leq H, w \leq W$, $p$ denotes the positive sample, $sli$ denotes the slice. Finally, the positive sample matrix $f_p \in \mathbb{R}^{c \times d \times h \times w}$ is generated by aggregating samples from $d$ slices.

#### 2) Constructing negative samples

In addition to providing consistent representations of the different parts of the pancreas, a promising model should also be able to distinguish the pancreas from the complex background. For this, we seek to improve the discriminability of the features in the border region. Specifically, we applied a random morphological dilation operator $D(\cdot)$ to the original pancreatic mask $Y_{ori}$ to obtain the dilated mask $Y_{dil}$, where the local neighborhood range $r$ of the dilation operation was obtained through uniform random sampling in the interval (10, 20). The original mask $Y_{ori}$ was subtracted from the dilated mask $Y_{dil}$ to obtain the background region $R$:

$$R = D\left(Y_{ori}, r\right) - Y_{ori}, \ r \sim U\left(10, 20\right). \tag{1}$$

The negative sample $f_n^{sli} \in \mathbb{R}^{c \times h \times w}$ is randomly extracted from surrounding non-pancreatic region $R$ slice by slice in $f_{proj} \in \mathbb{R}^{c \times D \times H \times W}$, where $n$ denotes the negative sample, $h \leq H, w \leq W$, $sli$ denotes the slice. Afterward, $d$ negative samples are concatenated into a matrix $f_n \in \mathbb{R}^{c \times d \times h \times w}$.

#### 3) Constructing contrastive pairs

For the positive and negative samples, we seek to maximize their intra-class consistency and inter-class separability. This enhances the integrity of the pancreatic features and the discriminability of the intra-pancreatic and extra-pancreatic features. Suppose there are $N$ pairs of positive and negative samples, respectively, i.e., $f_p, f_n$, wherein a total of $2N$ samples are used for the contrastive learning process. For two arbitrary samples $f_i, f_j$, we compute the cosine similarity as:

$$sim\left(f_i, f_j\right) = \frac{f_i^T f_j}{\|f_i\| \|f_j\|}. \tag{2}$$

where $i$ and $j$ denote different samples, respectively. This similarity is integrated into the InfoNCE [41] function:

$$\ell\left(f_i, f_j\right) = -log \frac{exp\left(sim\left(f_i, f_j\right) / \tau\right)}{\sum\limits_{i=1, i \neq j}^{2N} \mathbb{F}\left(f_i, f_j\right) \cdot exp\left(sim\left(f_i, f_j\right) / \tau\right)}. \tag{3}$$

where $\mathbb{F}(f_i, f_j)$ is an indicator, whose output is 0 if $f_i$ and $f_j$ belong to the same class, and 1 otherwise. $\tau$ denotes a temperature parameter, empirically set to 0.05. Let $C(N, 2)$ denotes the number of all sample pairs and it changes as the input changes, the contrastive loss $\mathcal{L}_{con}$ can be expressed as:

$$L_{con} = \sum_{i=1}^{2N} \sum_{j=i+1}^{2N} \frac{\left(1 - \mathbb{F}\left(f_i, f_j\right)\right) \cdot \ell\left(f_i, f_j\right)}{C\left(N, 2\right) \times 2}. \tag{4}$$

### B. Local-image-restoration Self-supervised Learning

Although the global-feature contrastive learning module can stress subtle feature discrepancies between the pancreas and its surrounding tissues, the intra-pancreatic anatomical features aren't still fully exploited. Furthermore, high uncertainty has been found to be associated with performance degradation [14-16]. Thus, we aim to exploit the anatomical context to further enhance the characterization of the sensitive high-uncertainty regions, and thus improve the segmentation robustness and generalization. Making use of the homogeneity and contiguity of the anatomical structures in 3D pancreas imaging, we propose a local-image-restoration self-supervised learning module. It accurately characterizes high-uncertainty regions and their surroundings based on effectively learning local-image-restoration, thus improving the segmentation task.

### 1) Identification of the high-uncertainty regions

We first identify the high-uncertainty regions which shall serve as the target for the subsequent local-image-restoration task. Specifically, $K$ data augmentation methods $A(\cdot)$ are applied to input $x$ to obtain different views: $\tilde{x}_k = A_k(x)$, where $A_k \in \{A_1, \cdots, A_K\}$ denotes a data augmentation technique, $k$ denotes a data augmentation method. Applying them to each sample $x$ individually, we obtain the following set of segmentation results from the global-feature self-supervised contrastive learning module $F_{GFS}$:

$$\tilde{y} = \left\{\tilde{y}_k\right\}_{k=1}^{K} = \left\{F_{GFS}\left(A_k\left(x\right)\right)\right\}_{k=1}^{K}. \quad (5)$$

Subsequently, we compute a pixel-level uncertainty maps $\tilde{y}_{unc}$ by combining the results $\tilde{y} = \{\tilde{y}_k\}_{k=1}^{K}$:

$$\tilde{y}_{unc} = \sqrt{\frac{1}{K}\sum_{k=1}^{K}\left(\tilde{y}_k - \frac{1}{K}\sum_{k=1}^{K}\tilde{y}_k\right)^2}. \quad (6)$$

The uncertainty map $\tilde{y}_{unc}$ is then binarized by a threshold $t$ to obtain the uncertainty mask $M$:

$$M_i = \begin{cases} 1, & if \; \tilde{y}_{unc}^i \geq t \\ 0, & if \; \tilde{y}_{unc}^i < t \end{cases}. \quad (7)$$

where $i$ denotes a pixel of the uncertainty map $\tilde{y}_{unc}$. To obtain training perturbation samples $I$ for the restoration self-supervised learning task, we introduce corruption patterns into the original appearance by randomly changing the truncation window of the CT radio-density values. The original image and the image with corrupted appearance are denoted by $I_{ori}$ and $I_{cor}$. High-uncertainty regions in $I_{ori}$ are replaced with the corresponding region in $I_{cor}$:

$$I = \left(1 - M\right) \cdot I_{ori} + M \cdot I_{cor}. \quad (8)$$

Besides, the initial probabilistic maps of the global-feature self-supervised contrastive learning module can be obtained as:

$$\hat{y} = \frac{1}{K}\sum_{k=1}^{K}\tilde{y}_k = \frac{1}{K}\sum_{k=1}^{K}F_{GFS}\left(A_k\left(x\right)\right). \quad (9)$$

Finally, the triples consisting of probabilistic maps $\hat{y}$, uncertainty maps $\tilde{y}_{unc}$ and perturbed samples $I$ serves as the input to local-image-restoration self-supervised learning module $F_{LIS}$, as Fig. 3 shows.

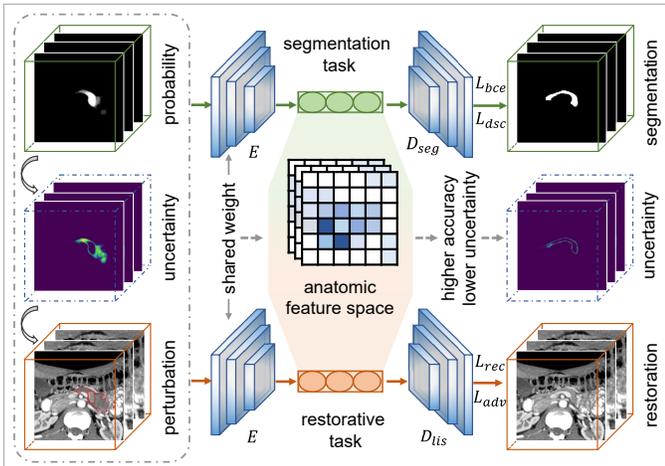

Fig. 3. A schematic illustration of the local-image-restoration self-supervised learning module.

### 2) Local-image-restoration self-supervised learning loss

This module consists of synergistic local-image-restoration self-supervised learning task and segmentation task, the loss $L_{lis}$ is thus obtained as the sum of the restoration self-supervised loss $L_{res}$ and the segmentation loss $L_{seg}$. Let $Y$ and $\hat{Y}$ denote the label and segmentation results, respectively. The segmentation loss $L_{seg}$ can in turn be expressed as the sum of the binary cross-entropy loss $L_{bce}$ and the Dice loss $L_{dsc}$:

$$L_{seg} = \frac{\sum\left(Y\log\left(\hat{Y}\right) + \left(1 - Y\right)\log\left(1 - \hat{Y}\right)\right)}{\sum Y} + 1 - \frac{2 \times \sum\left(Y\hat{Y}\right)}{\sum Y + \sum \hat{Y}}. \quad (10)$$

The self-supervised loss $L_{res}$ is the sum of the reconstruction loss $L_{rec}$ and the adversarial loss $L_{adv}$. Let $\hat{I}$ denotes the recovered image, $L_{rec}$ can be defined with the squared error $L_{sqe} = \left(I_{ori} - \hat{I}\right)^2$:

$$L_{rec} = \left(1 - \lambda\right)\frac{\sum\left(\left(1 - M\right) \cdot L_{sqe}\right)}{\sum\left(1 - M\right)} + \lambda\frac{\sum\left(M \cdot L_{sqe}\right)}{\sum M}. \quad (11)$$

where $\lambda$ is empirically set to 0.9. We also employ discriminator $D_{cls}$ and adversarial loss $L_{adv}$ to produce more realistic images:

$$L_{adv} = \mathbf{E}_{I \sim P\left(I_{ori}\right)}\log\left(D_{cls}\left(I_{ori}\right)\right) + \mathbf{E}_{\hat{I} \sim P\left(\hat{I}\right)}\log\left(1 - D_{cls}\left(\hat{I}\right)\right). \quad (12)$$

---

**Algorithm 1: Dual self-supervised learning framework**

**Training input:** training data $I_{ori}$, label $Y$.
**Training output:** global-feature self-supervised contrastive learning module $F_{GFS}$, local-image-restoration self-supervised learning module $F_{LIS}$ parameterized by shared encoder $E$ and decoder $D_{lis}$ and $D_{seg}$.
  1: **train** $F_{GFS}$ until converged
  2:   construct positive and negative samples  $f_p, f_n$. # Sec III.A.1-2
  3:   update parameters of $F_{GFS}$ using (4) and (10). # Sec III.A.3
  4: **train** $F_{LIS}$ until converged
  5:   obtain uncertainty mask $M$ using (5), (6) and (7). # Sec III.B.1
  6:   construct the input $I$ using (8) # Sec III. B.1
  7:   update parameters using (10), (11), and (12). # Sec III. B.2
**Testing input:** testing data $I_t$ from unseen source.
**Testing output:** segmentation results $\hat{Y}_t$.
  1: calculate probabilistic and uncertainty maps $\hat{y}, \tilde{y}_{unc} \leftarrow \mathcal{F}(I_t; F_{GFS})$.
  2: calculate segmentation results $\hat{Y}_t \leftarrow \mathcal{F}(I_t, \hat{y}, \tilde{y}_{unc}; F_{LIS}\{E, D_{seg}\})$.

---

### C. Learning, Backbone, and Implementation

For coarse stage, we applied the coarse model proposed in [42]. For fine stage, the proposed modules were independently trained due to memory limitations, as shown in Algorithm 1. For global-feature self-supervised learning, positive samples $f_p$ and negative samples $f_n$ were randomly collected from projected features $f_{proj}$ for training with the contrastive loss and the segmentation loss. Eight combinations of flips along three spatial directions were applied to obtain the uncertainty masks $M$, which was then used to construct perturbed samples $I$. To enhance the variability and avoid overfitting and early-learning difficulties, we adopted a progressive learning strategy. The threshold $t$ for identifying uncertainty maps was reduced linearly from 0.2 to 0.001, i.e., a larger $t$ was used in the early training to allow small region recovery and let the model converge quickly. A subsequent decrease enables the model to progressively focus on high-uncertainty regions. In the testing, the image-restoration branch $D_{lis}$ would be discarded, with $E$ and $D_{seg}$ retained for testing. The probabilistic maps $\hat{y}$ and uncertainty maps $\tilde{y}_{unc}$ were first obtained with $F_{GFS}$ and a fixed threshold $t$=0.01. Then, the samples were constructed based on $\hat{y}$, $\tilde{y}_{unc}$ and $I_t$ and fed into $F_{LIS}\{E, D_{seg}\}$.

We applied 3D U-Net [27] as the backbone for two proposed modules. It consists of four convolutional layers, where all batch normalization operators are replaced by group normalization ones. For the single-source generalization task, each dataset is only used in turn for model training while the remaining datasets from other sources are used independently for testing. We implemented our method using PyTorch and performed on an NVIDIA 3090 GPU. The Adam with a weight decay of 0.0005 was used for optimization with a learning rate of $1 \times 10^{-4}$ and an exponential decay $lr_{iter+1} = lr_{iter} \times ((1 - iter/max\_iter)^{0.9})$, where $iter$ denotes the current iterations and $max\_iter$ denotes the pre-set max iterations. The batch size is set to 1, the global-feature contrastive learning module and the local-image-restoration self-supervised learning module were trained for 500 and 200 epochs, respectively. Data augmentation was carried out through random rotations (-15°~15°), random flipping, and the addition of random noise. For the other two baseline models involved in the subsequent experiments, the patch size of the 3D patch method [27] is set to 64×120×120, the batch size is set to 2. The Adam optimizer a weight decay of 0.0001 is used with a learning rate of $1 \times 10^{-4}$ for training, and a learning rate dynamic decay strategy is also employed. The 2.5D XYZ fusion method [30] begins by importing pre-training parameters from the PascalVOC dataset [43], followed by training with an SGD optimizer with a momentum of 0.99 and a weight decay of 0.0005 utilizing a batch size of 1 and a learning rate of $1 \times 10^{-5}$.

## TABLE I
### Key Information of Three Pancreatic CT Imaging Datasets.

| Data | Institution | Cases | Tumors | Slices | Spacing (mm) | Thickness (mm) |
|------|-------------|-------|--------|--------|--------------|----------------|
| NIH | The national institutes of health clinical center | 82 | **No** | 181-466 | 0.66-0.98 | 1.50-2.50 |
| MSD | Memorial Sloan Kettering cancer center | 281 | Yes | 37-751 | 0.60-0.97 | 0.70-7.50 |
| Private | Renji Hospital of Shanghai Jiao Tong University School of Medicine | 104 | Yes | 152-341 | 0.55-0.88 | 0.79-1.25 |

## IV. Results

### A. Dataset Description

We evaluated the performance of our method with three different pancreatic CT imaging datasets, namely the NIH dataset [44], the MSD dataset [45] and a private dataset (denoted as 'Private'). Table I summarizes key pieces of information: the medical institution where the dataset was collected, the case count, whether malignant cases, the number of slices for each case, pixel spacing and the slice thicknesses. The NIH and MSD datasets are widely-used public ones, while the private dataset was locally collected from Renji Hospital of Shanghai Jiao Tong University School of Medicine. This study was approved by Shanghai Jiao Tong University School of Medicine, Renji Hospital Ethics Committee (IRB No. RA-2021-094). The MSD and the Private datasets are collected from pancreatic cancer patients while the NIH dataset has normal subjects only. The Private dataset was comprised of patients with pancreatic ductal adenocarcinoma (PDA), while the MSD dataset was sampled from patients undergoing resection of pancreatic masses, including intraductal papillary mucinous neoplasm (IPMN), pancreatic neuroendocrine tumor (PNET), and PDA. The MSD samples were resampled to 1.5×0.8×0.8 to account for the large thickness and fit the 3D model. The CT radio-density values were truncated to the range of [-100, 240] Hounsfield units (HU) and normalized to [0,1].

### B. Evaluation Metrics

We employ the Dice similarity coefficient (DSC) as evaluation metrics. The DSC metric assesses the similarity between the ground-truth and the segmentation result by computing the proportion of overlap among the two regions. A larger DSC value indicates a more accurate segmentation result.

### C. Quantitative and Qualitative Results

#### 1) Analysis of the baseline methods

The existing pancreas segmentation methods can be mainly categorized into one-stage and two-stage schemes. The state-of-the-art (SOTA) one-stage method is the 3D patch method of Zhang et al. [27], while the SOTA two-stage methods are the 3D coarse-to-fine (3D-C2F) method [4] and the 2.5D XYZ fusion (XYZ-F) method [30]. We evaluated the performance of these three baseline schemes for the four-fold cross-validation (CV) and generalization (GL) tasks. As shown in Table II and Table III, the 3D-C2F scheme shows the best performance on both tasks. So, this scheme was adopted as the powerful baseline scheme for subsequent experiments. Furthermore, the large gap between the CV and GL results further shows the large effects of the imaging variabilities on the model robustness, demonstrating the necessity and challenge of single-source generalization task.

## TABLE II
### Four-fold Cross Validation for Three Baseline Methods (DSC %)

| Data | NIH | MSD | Private | Mean |
|------|-----|-----|---------|------|
| 3D patch | **85.23**[1] | 79.70[2] | 77.39 | 80.77 |
| XYZ-F | 84.53[3] | 82.74[4] | **82.96** | 83.40 |
| 3D-C2F | 84.64 | **83.55** | 82.92 | **83.70** |

[1-4]: The results are reported in [27], [35], [31], and [34], respectively.

## TABLE III
### Generalization Performance for Three Baseline Methods (DSC %)

| Train | NIH | | MSD | | Private | | Mean |
|-------|-----|-----|-----|-----|---------|-----|------|
| Test | MSD | Private | NIH | Private | NIH | MSD | |
| 3D patch | 49.40 | 62.92 | 67.67 | 74.69 | 56.11 | 68.08 | 63.14 |
| XYZ-F | 57.52 | 65.84 | 68.08 | 78.69 | **64.58** | 67.82 | 67.08 |
| 3D-C2F | **61.44** | **69.08** | **77.57** | **82.66** | 59.30 | **68.52** | **69.76** |

#### 2) Analysis of the ablation study

We conducted ablation experiments to investigate the effectiveness of each module of our proposed method on the single-source generalization task. As stated earlier, the standard 3D coarse-to-fine (3D-C2F) model [4] was adopted as our baseline (BL). The global-feature self-supervised learning (GFS) module and the local-image-restoration self-supervised learning (LIS) module were then independently added to the BL, to examine the individual contributions. Finally, both the GFS and LIS modules were added together to the BL to form our final model (Ours) to confirm the compatibility and joint performance gains of the two modules. Table IV and Fig. 4 demonstrate a continuous performance improvement as the proposed modules are added -one at a time- to the baseline model. Besides, we report in Table V the average number of pixels with an uncertainty greater than 0.01 and show sample results in Fig. 5. Clearly, our method significantly reduced the segmentation uncertainty and successfully improved the segmentation performance in the high-uncertainty regions. This comprehensive ablation study illustrates the effectiveness of each of the two main components of our method.

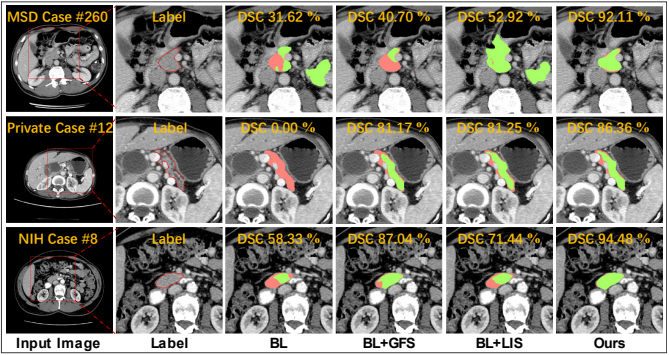

Fig. 4. Example results for the ablation study. The green regions denote the segmentation results, while the red regions denote the ground-truth.

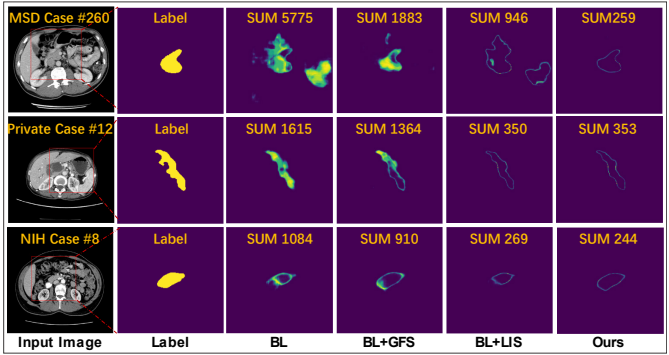

Fig. 5. Uncertainty maps in the ablation experiments. The shown sum indicates the number of all pixels with uncertainty exceeding 0.01.

TABLE IV
ABLATION STUDY OF OUR METHOD (GFS: GLOBAL FEATURE SELF-SUPERVISED, LIS: LOCAL IMAGE SELF-SUPERVISED, DSC %).

| Train | NIH | | MSD | | Private | | Mean |
|---|---|---|---|---|---|---|---|
| Test | MSD | Private | NIH | Private | NIH | MSD | |
| Baseline (BL) | 61.44 | 69.08 | **77.57** | 82.66 | 59.30 | 68.52 | 69.76 |
| BL + GFS | 63.81 | 71.59 | 77.51 | **83.92** | 60.59 | 69.14 | 71.09 |
| BL + LIS | 65.81 | 72.97 | 76.72 | 82.85 | 61.63 | 69.76 | 71.62 |
| **Ours** | **66.73** | **73.85** | 76.71 | 83.50 | **65.03** | **70.08** | **72.65** |

TABLE V
THE MEAN NUMBER OF PIXELS WITH UNCERTAINTY GREATER THAN 0.01 IN DIFFERENT MODELS.

| Train | NIH | | MSD | | Private | | Mean |
|---|---|---|---|---|---|---|---|
| Test | MSD | Private | NIH | Private | NIH | MSD | |
| Baseline (BL) | 81867 | 116160 | 130002 | 111935 | 171985 | 95474 | 117903 |
| BL + GFS | 83457 | 114557 | 136895 | 104783 | 136126 | 82187 | 109667 |
| BL + LIS | **21521** | **28501** | **44621** | 40118 | 43698 | 23890 | 33724 |
| **Ours** | 21967 | 28935 | 48935 | **38189** | **38430** | **21998** | **33075** |

*a) Ablation study results for the global-feature self-supervised learning module*: We could observe the effectiveness of the proposed global feature self-supervised (GFS) module from two sets of experiments, namely, BL versus BL+GFS and BL+LIS versus Ours. The GFS brings average performance gains of 1.33% and 1.03% for the BL and BL+LIS models, respectively (Table IV). In other words, the GFS module significantly improves the baseline performance, and contributes clear performance gains together with the LIS module as well. As shown in Fig. 4, the BL+GFS improves certain BL results that were initially unreliable due to the influence of the surrounding tissues. The BL+GFS not only eliminates over-segmentation (the first row in Fig. 4 and Fig. 5), but also alleviates under-segmentation (the second row in Fig. 4 and Fig. 5). This makes results more focused on the true pancreatic regions. We used t-SNE [46] to visualize the embeddings of pancreatic and non-pancreatic regions in Fig. 6, demonstrating successful feature boundary expansion by GFS.

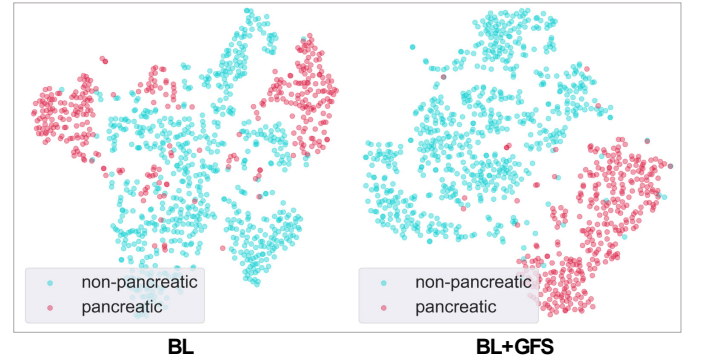

Fig. 6. Visualization of the learned feature embedding of pancreatic and non-pancreatic regions with different methods.

*b) Ablation study results for the local-image-restoration self-supervised learning module*: We can also observe the effectiveness of the proposed local-image-restoration self-supervised (LIS) module from two experiments, namely, BL versus BL+LIS and BL+GFS versus Ours. We note that the LIS module contributes average performance gains of 1.86% and 1.56%, respectively (Table IV). Besides, with the rich anatomical context exploited by the pixel-level image-restoration module, the segmentation stability in the high-uncertainty regions is greatly enhanced. This shows excellent robustness in most regions, with only few negligible segmentation inconsistencies at the pancreatic boundaries (Table V and Fig. 5).

*c) Ablation study results for the dual self-supervised learning framework*: From the results above, it can be observed that the LIS module can use the BL segmentation confidence as a cue to explore the anatomical context and hence boost the segmentation results (See the second and third rows in Fig. 4 and Fig. 5). However, excessive interference from the tissues surrounding the pancreas can cause the LIS module to exhibit high segmentation uncertainty (See the sixth column and first row in Fig. 4 and Fig. 5). Fortunately, this challenge can be greatly alleviated by the GFS module with the elimination of surrounding disturbances. Indeed, the LIS and GFS modules complement each other and achieve more stable segmentation results (See the first row in Fig. 4 and in Fig. 5).

### 3) Analysis of the performance comparison with other generalization methods

We compared the performance outcomes of several SOTA single-source generalization methods, including the methods of BigAug [8], Cutout [36], DAE [39], MixStyle [37] and RSC [38]. To ensure a fair comparison, all of these methods were implemented with the same baseline architecture as ours, namely, the 3D-C2F architecture [4]. As shown in Table VI, the performance of each of the RSC, Cutout and BigAug methods is restricted by the small amount of the single-source training data. Indeed, these methods demonstrate limited performance advantages over the baseline model. The MixStyle, DAE, and RandConv methods, while showing promising performance, have widely varying generalization performance for datasets from different sources, so the average result decreases. Our method exhibits higher overall performance and enhanced robustness on the generalization task for three datasets. Besides, Fig. 7 shows the generalization performance of these methods on three representative cases from the three datasets. Our method can accurately preserve the complete pancreas shape and eliminate redundant parts for the segmentation on unseen datasets, whereas the other methods sometimes failed to do so.

#### TABLE VI
COMPARISON OF OUR METHOD WITH OTHER SINGLE-SOURCE GENERALIZATION METHODS (DSC %).

| Train | NIH | | MSD | | Private | | Mean |
|---|---|---|---|---|---|---|---|
| Test | MSD | Private | NIH | Private | NIH | MSD | |
| RSC [38] | 61.1 | 64.85 | 67.81 | 80.21 | 50.47 | 58.71 | 63.85 |
| Cutout [36] | 60.22 | 68.86 | 77.76 | 83.3 | 60.22 | 68.54 | 69.81 |
| Big Aug [8] | 62.69 | 71.12 | 77.36 | 83.28 | 59.08 | 67.86 | 70.23 |
| MixStyle [37] | 62.07 | 70.45 | 78.19 | 83.28 | 61.90 | 69.25 | 70.85 |
| DAE [39] | 63.34 | 69.18 | **79.07** | **83.71** | 63.70 | 68.97 | 71.32 |
| **Ours** | **66.73** | **73.85** | 76.71 | 83.50 | **65.03** | **70.08** | **72.65** |

#### TABLE VII
FOUR-FOLD CROSS-VALIDATION RESULTS OF THE BASELINE AND OUR DUAL SELF-SUPERVISED LEARNING METHOD (DSC %).

| Train | NIH | MSD | Private | Mean |
|---|---|---|---|---|
| Baseline | 84.64 | 83.55 | 82.92 | 83.70 |
| **Our Method** | **85.64** | **84.65** | **84.09** | **84.79** |

#### 4) Analysis of the impact on the cross-validation tasks

Numerous promising approaches have been proposed to boost the performance on unseen data, but the impact of these approaches for local data has not been adequately addressed. Ideally, a segmentation model should not only be applicable to unseen data (i.e., generalization task), but also should not compromise performance on local in-distribution data (i.e., cross-validation task). Thus, to investigate whether our dual self-supervised learning framework performs well on local in-distribution data, we carried out a four-fold cross-validation scheme. We report the results in Table VII. Obviously, it yields performance gains on the cross-validation task for the three datasets. So, our method is beneficial (rather than detrimental) to the segmentation performance on local in-distribution data.

#### 5) Analysis of the model hyperparameters

To investigate the effect of the uncertainty threshold $t$ on the generalization performance, we conducted an ablation study for this. We set the threshold to four different orders of magnitude and got stable results with fluctuations not exceeding 0.42% in Table VIII. Besides, to investigate the effect of the number $K$ of data augmentation procedures used in constructing uncertainty maps, we also performed an ablation study of $K$. Table IX shows that close results are obtained even if only four data augmentation procedures are used. In fact, larger $K$ values lead to more reliable uncertainty maps. In general, our model can achieve stable results under different qualities of the uncertainty maps. These results show that our method effectively exploits cues of uncertainty and anatomical context, and hence adequately focuses on the surroundings of the high-uncertainty regions (rather than being limited to a fixed region). Overall, our method is not sensitive to either $t$ or $K$, and can exhibit excellent performance under different hyperparameters.

#### TABLE VIII
PERFORMANCE WITH DIFFERENT THRESHOLDS. (DSC %).

| $t$ | NIH | | MSD | | Private | | Mean |
|---|---|---|---|---|---|---|---|
| | MSD | Private | NIH | Private | NIH | MSD | |
| 0.1 | 65.00 | 72.38 | **77.42** | **83.67** | **65.23** | 69.68 | 72.23 |
| **0.01**[*] | 66.73 | 73.85 | 76.71 | 83.50 | 65.03 | **70.08** | **72.65** |
| 0.001 | 67.43 | 74.62 | 76.11 | 83.34 | 63.65 | 70.00 | 72.52 |
| 0.0001 | **67.47** | **74.79** | 75.73 | 83.20 | 62.94 | 69.91 | 72.34 |

* The value used in this work.

#### TABLE IX
PERFORMANCE WITH DIFFERENT AUGMENTATION METHODS. (DSC %).

| $K$ | NIH | | MSD | | Private | | Mean |
|---|---|---|---|---|---|---|---|
| | MSD | Private | NIH | Private | NIH | MSD | |
| 3 | 64.86 | 72.05 | 76.17 | 83.49 | 65.16 | 69.55 | 71.88 |
| 4 | 65.60 | 73.01 | **77.21** | **83.59** | 64.76 | 69.75 | 72.32 |
| 5 | 66.00 | 73.76 | 76.88 | 83.48 | **65.25** | 69.67 | 72.50 |
| 6 | 66.33 | 73.45 | 76.65 | 83.42 | 64.49 | 69.91 | 72.37 |
| 7 | **66.99** | **73.89** | 76.70 | 83.45 | 64.04 | **70.11** | 72.53 |
| **8**[*] | 66.73 | 73.85 | 76.71 | 83.50 | 65.03 | 70.08 | **72.65** |

* The value used in this work.

#### 6) Analysis of the computational efficiency

To highlight the computational costs of different pancreas segmentation methods, we present in Table X their parameter counts and the inference time. Since none of these methods explored the generalization performance, we investigate the computational efficiency with a four-fold cross-validation on the NIH dataset. The 3D patch method [27] has the smallest parameter count because only one model should be trained, but the inference time is significantly increased due to the sliding windows for testing patches. The XYZ-fusion approach tends to have more model parameters and a longer inference time due to the independent training and testing from three different views. The method in [33] has been carefully designed to be a lightweight one, and therefore it has the shortest inference time. Although the computational cost of our method is increased due to the two-stage process and the uncertainty estimation, the parameters and the inference time are still smaller than those of most methods. In conclusion, our method achieves encouraging performance with the second lowest number of parameters and inference time, this demonstrates the efficiency of our approach.

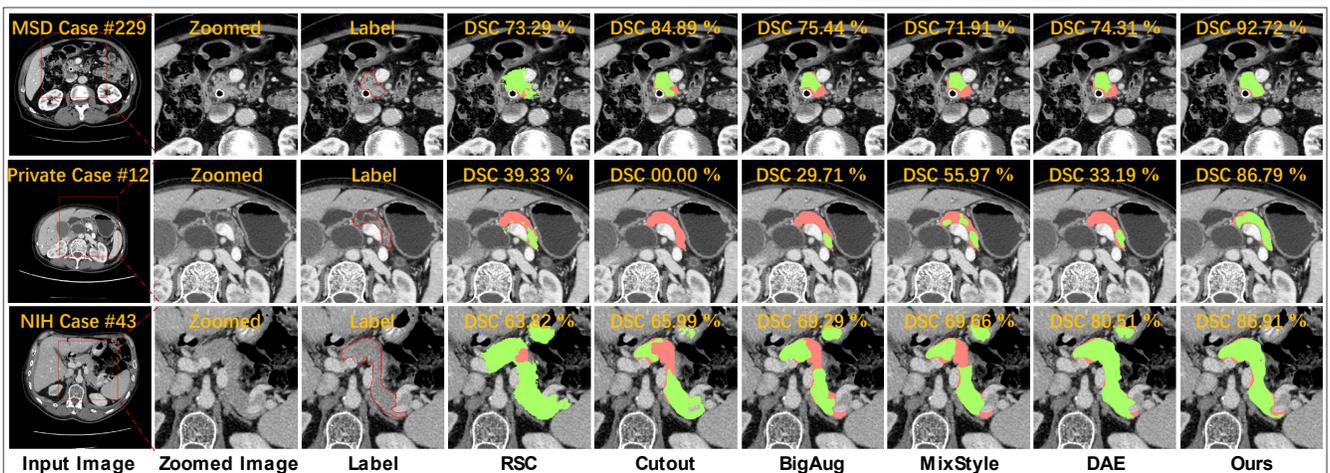

Fig. 7. Segmentation results for different methods. The green regions denote the results, while the red regions denote the ground-truth.

TABLE X
MODEL PARAMETER COUNTS AND INFERENCE TIMES.

| Methods | DSC* | Parameters (×10⁶) | Time (s/case) | Method | Devices |
|---|---|---|---|---|---|
| Zhang et al. [34] | 84.47 | / | 180-300 | XYZ-Fusion | / |
| Xie et al. [31] | 84.53 | 256.11 | 78.0 | XYZ-Fusion | Titan-X Pascal |
| Zhang et al. [33] | 84.90 | 25.13 | 14.0 | XYZ-Fusion | GTX 1080Ti |
| Li et al. [5] | 85.35 | 75 | 134.3 | XYZ-Fusion | GTX 2080Ti |
| Hu et al. [32] | 85.49 | 65.16 | 77.3 | XYZ-Fusion | GTX 2080Ti |
| Zhang et al. [27] | 86.06 | 18.8 | 92.5 | 3D Patch | GTX 3090 |
| Baseline | 84.64 | 37.91 | 14.8 | 3D C2F | GTX 3090 |
| Ours | 85.64 | 56.88 | 21.9 | 3D C2F | GTX 3090 |

* Because none of these pancreas segmentation methods provide generalization results, DSCs are obtained by four-fold cross-validation on NIH dataset.

## V. DISCUSSION

This work takes the first step towards addressing the challenging task of generalizable pancreas segmentation with single-source data. A common practice is to collect high-quality labeled data from multiple medical institutions to capture as much of the potential imaging variations as possible. However, inadequate data sources exposed the model to few instances of imaging variations, yielding poor performance on unseen data. Thus, this study extensively explored the remarkable role of the rich anatomical information in enhancing the characterization of the high-uncertainty regions, and hence improving the generalization performance. Notably, our method is model-independent and can be integrated into other models to boost the segmentation performance in the high-uncertainty regions.

Recently, many methods have achieved promising pancreas segmentation results on local in-distribution data. Despite the importance of transferring algorithms among datasets has attracted attention [31], no generalizable pancreas segmentation methods is proposed due to the huge challenges of this task. The pancreas segmentation work that is the most relevant to ours is that of Hu et al. [7], who constructed a pseudo-generalization environment. Specifically, a Gaussian mixture model was used to aggregate the MSD samples into two groups: a source dataset (184 cases) and a target dataset (97 cases). The two datasets were used to simulate the generalization scenario, and achieve a DSC value of 72.85% on the target dataset. In this work, we employed our private dataset (104 cases) as the source dataset and achieved a DSC value of 70.08% on the whole MSD dataset. Though the results in [7] is remarkable, it is still essentially an exploration of the segmentation on local in-distribution data, while our method employed a smaller training set and a realistic generalization scenario to yield a DSC result that is only 2.77% less. This demonstrates the effectiveness of our approach.

We follow the assumption that higher-uncertainty regions often show degraded segmentation performance compared to lower-uncertainty regions [14-16]. This assumption also holds in our work, as shown in Fig. 4 and Fig. 5, where the high-uncertainty regions have higher probabilities of segmentation errors. Thanks to our dual self-supervised learning strategy, we not only reduced the interference of the tissues surrounding the pancreas, but we also effectively exploited the local anatomical context to enhance the characterization of the high-uncertainty regions. The results also show that our method successfully improves the segmentation accuracy (See the last column of Fig. 4) and reduces the corresponding uncertainty (See the last column of Fig. 5). In this work, we utilized a simple but efficient strategy based on a shared encoder to synergize the local-image-restoration self-supervised learning task with the segmentation task. Thus, it would be interesting to further explore how to facilitate information transfer and fusion between these two tasks to better exploit the anatomical context.

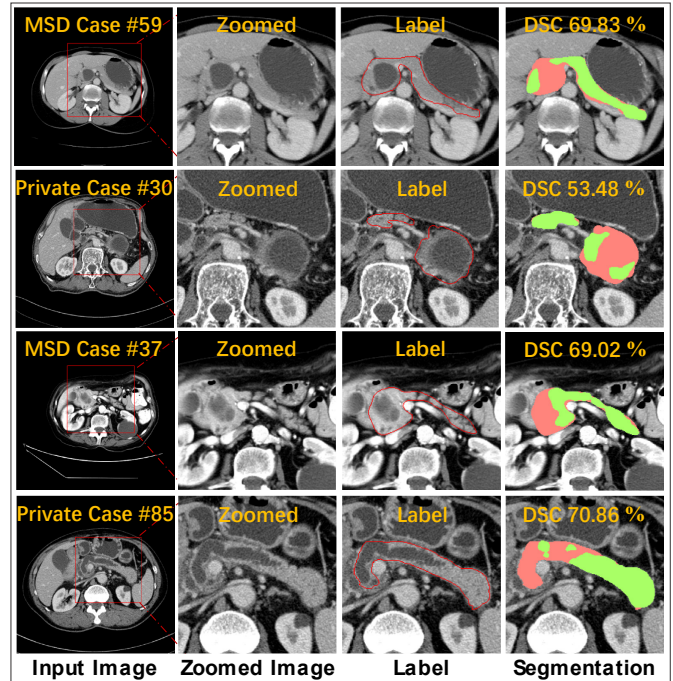

Fig. 8. Failure cases for models trained with the NIH dataset and tested on the MSD and Private datasets. The green regions denote the segmentation results, while the red regions denote the ground-truth.

This study utilized three datasets encompass distinct patient groups. The NIH dataset only comprises normal subjects, while the Private dataset is comprised of PDA subjects. The MSD dataset containing three types of lesions (IPMN, PNET and PDA) exhibited artifacts due to the presence of biliary stents in approximately 74 cases [47]. These factors exhibit varying pancreatic imaging characteristics, thus presenting additional intricacies to generalization task. Firstly, tumor invasion led to diverse pancreatic appearances, making the generalization across normal and tumor samples more challenging. For instance, models trained on NIH data may mistakenly classify tumor-affected pancreatic tissues as background, causing under-segmentation errors (Fig. 8). Secondly, distinct tumor types yield varying radiographic findings, potentially impacting the generalization across samples with different lesions. For example, models trained on the MSD dataset with IPMN, PNET, and PDA samples could generalize well to the Private dataset with only PDA samples, but the generalization performance from the Private to the MSD deteriorated significantly due to the absence of IPMN and PNET samples in the training set (Table IV). Finally, we observed that the LIS module trained on the MSD dataset performed poorly on NIH dataset (Table IV), mainly due to data differences caused by biliary stents and diverse types of pancreatic lesions. These factors increase uncertainty in the corresponding regions, leading the LIS module to focus on these regions during training. But the absence of stents and heterogeneous lesions in NIH and Private datasets hinders the expected performance of the LIS module. Moreover, our method demonstrates good performance not only when trained on the NIH or Private datasets (Table IV) but also in all intra-domain cross-validation tasks (Table VII). This finding highlights the impact of the data disparity on generalization performance. Overall, addressing these challenges is crucial in advancing prospective applications in diverse clinical realms. Future work will accentuate the challenges associated with samples exhibiting different health conditions, involving the curation of multi-pathology pancreatic imaging data and the development of methodologies to mitigate the impact of heterogeneous patient populations on generalization performance.

## VI. Conclusion

A challenging problem that arises in generalization task is the reduced adaptability and stability of models trained with limited data to unseen data with wide imaging variabilities. We propose a dual self-supervised learning framework, through which we aim to systematically explore the anatomical information from intra-pancreatic and extra-pancreatic regions, to enhance the characterization of low-generalization high-uncertainty regions. Our method has two complementary and cooperative modules. The global-feature self-supervised contrastive learning module is applied to eliminate interference from the tissues surrounding the pancreas. Further segmentation enhancement is achieved with local-image-restoration self-supervised learning module, which exploits the anatomical context of the high-uncertainty regions to achieve more accurate characterization of these regions and hence attain better segmentation performance. Our work is the first attempt (as far as we know) to address the issue of single-source generalizable pancreas generalization. We provide a comprehensive evaluation of multiple pancreas segmentation and generalization methods, this shall hopefully stimulate further research along these meaningful lines.